\documentclass{article}

\usepackage[preprint]{neurips_2024}

\usepackage{pifont}

\usepackage[utf8]{inputenc}
\usepackage[T1]{fontenc}
\usepackage{hyperref}
\usepackage{url}
\usepackage{booktabs}
\usepackage{amsfonts}
\usepackage{xspace}
\usepackage{nicefrac}
\usepackage{microtype}
\usepackage{xcolor}
\usepackage{enumitem}
\usepackage{makecell}
\usepackage{multirow}
\usepackage{placeins} 

\usepackage{amsmath}
\usepackage{graphicx}

\newcommand{\trace}{\textsc{TRACE}\xspace}

\begin{document}

\title{TRACE: Training Reasoning Agents for Causal Exploration with Synthesized Rewards}

\author{
Rui Sun \quad Zhan Shi \quad Bing He \\
Independent Researchers
}

\maketitle

\begin{abstract}
Reinforcement learning with verifiable rewards (RLVR) has advanced language-model reasoning in domains such as mathematics and code, where objective answers are inexpensive to check. 
Diagnostic reasoning over complex data lacks this advantage: establishing the true cause of an anomaly often requires costly expert investigation and may remain ambiguous after the fact. 
We ask whether this asymmetry of verification can instead be engineered. 
We sample an intervention, inject it into a controlled simulator, and generate the observations it would produce. 
The hidden intervention provides an oracle label and objective reward, while the agent must still investigate noisy, confounded, and distributed evidence.

We instantiate this approach in \trace{}, a digital-advertising diagnostic environment with 12 root causes and fine-grained segment attribution. 
Agents investigate each episode using Python and SQL and must identify both the root cause and, when applicable, the affected segment assignment. 
On a held-out 235-episode test set, the strongest prompted baseline, Claude Opus~5, reaches $0.686$ FullAttr@1. 
Supervised fine-tuning raises Qwen3.5-35B-A3B from $0.159$ to $0.637$, and subsequent RL with synthesized rewards reaches $0.757$, outperforming all evaluated prompted baselines, including frontier closed-source models and a prompted Qwen3.5-122B-A10B model. 
The resulting policy also uses substantially fewer tool calls than the prompted 35B base.
These results provide evidence that access to a scalable, objective training signal can be a more important constraint than model scale alone. 
More broadly, simulation-based verification can make otherwise ambiguous diagnostic reasoning tasks amenable to scalable reinforcement learning.
\end{abstract}

\section{Introduction}

Reinforcement learning with verifiable rewards (RLVR) has driven recent progress in LLM reasoning. 
Work on mathematical reasoning~\citep{guo2025deepseek, shao2024deepseekmath} and coding agents~\citep{wei2026swe,liu2025deepswe} shows that policies optimized against objective verifiers can acquire multi-step reasoning and tool-use capabilities. 
These domains provide a natural \emph{asymmetry of verification}: checking a solution against a known answer or test suite is considerably easier than producing the solution itself~\citep{wei2025asymmetry}. 
RLVR turns this asymmetry into a scalable training signal.


Many practical reasoning tasks lack such verifiers. 
Diagnostic reasoning over complex data is a representative example: determining why a system changed can require expensive expert investigation, and the resulting attribution may remain uncertain even after the fact. 
Multiple changes can occur concurrently, observations are noisy, and plausible causes may produce similar signatures.
LLM-as-judge approaches~\citep{zheng2023judging,bai2022constitutional,lee2024rlaif} do not eliminate this bottleneck because the target itself remains ambiguous, while learned judges can introduce additional opportunities for reward hacking~\citep{gao2023scaling,
skalse2022defining}. 
The central obstacle to applying RLVR in these settings is therefore access to scalable, objective ground truth.


In this paper, we ask whether the asymmetry of verification can be \emph{engineered}.
Rather than observing data and attempting to determine an unknown cause, we first sample an intervention, inject it into a controlled simulator, and generate the observations that the intervention would produce. 
The hidden intervention is retained as an oracle label, making the agent's final attribution inexpensive and deterministic to verify. 
The agent does not observe this label: it must still investigate noisy, confounded, and distributed evidence to recover the cause. 
Simulation therefore separates the difficulty of solving a diagnostic task from the difficulty of verifying its answer.



We instantiate this approach in \trace{} (Training Reasoning Agents for Causal Exploration), a simulated diagnostic environment for digital advertising. 
Each \trace{} instance presents an agent with a campaign-performance anomaly and a multi-table database. 
The agent investigates through Python and SQL, iterating between hypotheses and evidence, and ultimately attributes the anomaly to one of the predefined cause types while ruling out realistic confounders.
\trace{} builds a \emph{simulator--oracle--RL pipeline}: 
the simulator injects a hidden intervention, the oracle verifies that the agent-visible data contains sufficient evidence to recover it, and RL uses the resulting hidden label as a synthesized reward. 
We refer to this setting as \emph{RL with synthesized rewards}, where hidden intervention labels are converted into objective reward signals.

We evaluate on a held-out 235-episode test set enriched for segment-specific attribution. 
The strongest prompted baseline, Claude Opus~5, reaches $0.686$ FullAttr@1. 
On Qwen3.5-35B-A3B, SFT raises FullAttr@1 from $0.159$ to $0.637$, and subsequent RL with synthesized rewards improves it to $0.757$, surpassing every evaluated prompted baseline. 
The post-trained 35B model also substantially outperforms the prompted Qwen3.5-122B-A10B model, which reaches $0.283$. 
This result provides evidence that, for this diagnostic setting, the binding constraint is access to an effective post-training signal---including a scalable, objective reward---rather than model scale alone. 
The best trained policy also averages $11.73$ tool calls per
trajectory, compared with $22.05$ for the prompted 35B base.

Our contributions are threefold.
First, we introduce a simulator--oracle--RL methodology for synthesizing objective reward in domains where natural verifiers are scarce.
The key idea is to generate tasks from controlled interventions, so that the same process that creates a difficult reasoning problem also provides an objective answer.
Second, we instantiate this methodology in \trace{}, a tool-using diagnostic environment with configurable causes, realistic confounders, and multi-table evidence. 
\trace{} supports held-out evaluation and post-training without human attribution labels or an LLM judge. 
Third, we show that SFT followed by RL with \trace{}'s synthesized rewards substantially improves an open-weight 35B reasoning agent, enabling it to outperform frontier closed-source models on diagnostic attribution tasks. 
We also conduct ablation experiments to study the roles of supervised initialization and RL reward design in producing these gains.
Together, these results show how simulation-based verification can make otherwise ambiguous diagnostic reasoning tasks amenable to scalable reinforcement learning.



\subsection{Related Work}\label{sec:related}

\paragraph{RLVR and reasoning training.}
RLVR methods optimize policies against verifiable outcomes, including
mathematical answer checkers~\citep{shao2024deepseekmath,guo2025deepseek},
formal proof validators~\citep{kim2026process}, and compilation or execution tests for code~\citep{wei2026swe,liu2025deepswe}.
Recent work has broadened RLVR by improving the underlying policy optimization algorithms~\citep{shao2024deepseekmath,yu2026dapo,liu2025understanding,zheng2025group} and by extending the training recipe beyond math and code, for example to medical question answering~\citep{chen2025towards}.
However, these advances still assume that verifiable targets already exist.
Our work instead studies how to construct such targets for domains where
they are not naturally available.

\paragraph{Synthetic data, simulators, and reward synthesis.}
Synthetic generation has been widely used to create instruction-following
data and reasoning trajectories for supervised fine-tuning
\citep{wang2023self,luo2025wizardmath,yu2024metamath}.
In these settings, the generator primarily supplies examples to imitate.
A related line of work develops simulated and interactive environments for evaluating tool-using agents in science, web, and machine-learning tasks~\citep{wang2022scienceworld,zhou2024webarena,huang2024mlagentbench}. These environments use interaction to test whether agents can complete tasks.
\trace{} uses simulation differently: it samples a hidden root cause,
generates the diagnostic data it would produce, and uses the sampled cause as the oracle label for reward calculation.

\paragraph{Tool-using data-analysis and diagnostic agents.}
A growing line of work evaluates whether LLM agents can analyze structured data with external tools.
Text-to-SQL benchmarks provide an early testbed for this setting by
asking models to translate natural-language questions into executable
database queries
\citep{yu2018spider,zhong2017seq2sql,li2023can,chang2023dr}.
More recent data-analysis benchmarks move toward multi-step tool use over tables, files, and code execution \citep{huang2024mlagentbench,zhang2026datascibench}.
These tasks capture important components of diagnostic investigation, but they generally score query correctness, code outputs, or final reports rather than root-cause attribution under controlled confounders.
\trace{} targets this gap by constructing diagnostic tasks with simulated interventions, agent-visible evidence, and verifiable rewards.

\paragraph{Causal reasoning and root-cause attribution.}
A parallel line of work evaluates whether LLMs can answer causal questions, reason over causal graphs, or recover causal structure from data \citep{jin2023cladder,jin2024can,kiciman2024causal,wang2024causalbench}.
These benchmarks probe causal knowledge and formal causal reasoning, often through static questions, symbolic graphs, or fixed datasets. \trace{} studies a different setting: interactive root-cause attribution in a simulated environment where the true cause is known by construction. 
The agent must gather evidence through SQL and Python, distinguish the sampled cause from realistic confounders, and produce an attribution that can be verified against ground-truth labels.

\section{The \trace Environment}\label{sec:env}
\begin{figure*}[h]
    \centering
    \includegraphics[width=\textwidth]{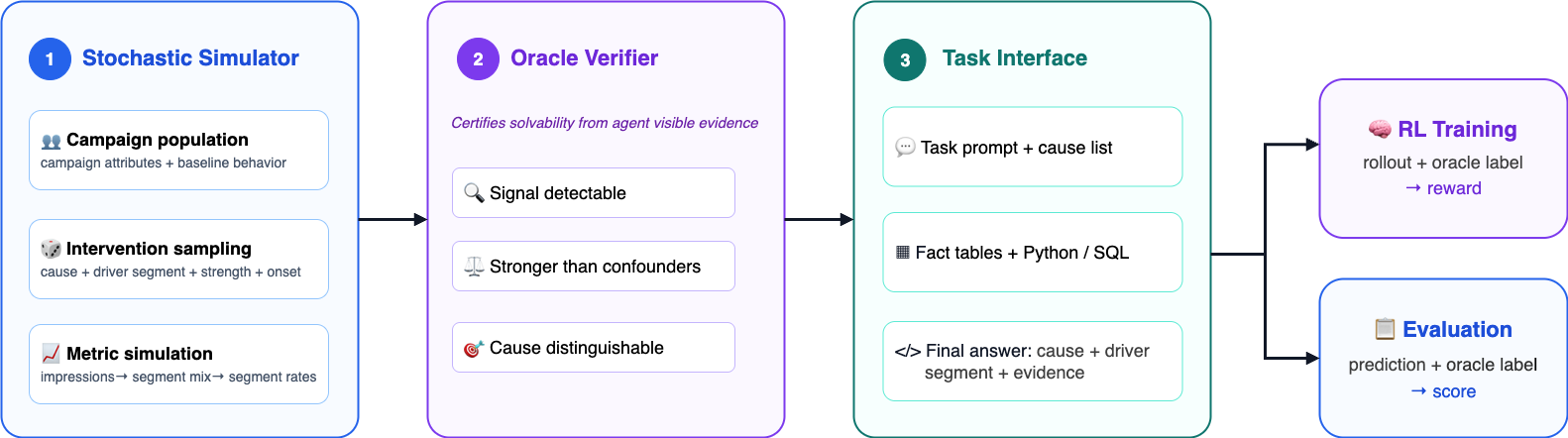}
    \caption{
    Overview of the \trace{} workflow.
    }
    \label{fig:trace_overview}
\end{figure*}

This section describes \trace{}, a digital advertising environment that implements the simulator--oracle components of our methodology.
\trace has three parts: a stochastic simulator that generates multi-table advertising data with known injected causes, a task interface through which an agent investigates each generated diagnostic instance, and an oracle verifier that certifies solvability and produces reward labels.
Figure~\ref{fig:trace_overview} summarizes the overall workflow.
We refer to each generated diagnostic instance as an \emph{episode}: the observable history of a single campaign over a fixed time window, together with a hidden injected intervention that defines the ground-truth cause.

\subsection{Preliminaries: Campaign Metrics and Diagnosis}\label{sec:preliminaries}

Digital advertising campaigns serve ads to users through online auctions: when a user loads a page with an ad slot, advertisers bid for the slot, and the winning ad is shown as an \emph{impression}. 
Under the commonly adopted pay-per-click model, the advertiser is charged only when the user \emph{clicks}, and a click that leads to a purchase of the advertised product is recorded as an \emph{order}, or \emph{conversion}.
Campaign performance is usually summarized by metrics such as impressions, click-through rate (CTR, clicks/impressions), conversion rate (CVR, orders/clicks), and cost-per-click (CPC, total advertiser spend/clicks).\footnote{
In general second-price auction mechanisms, an advertiser's paid price depends on its competing bids, not the advertiser's own. We model the resulting gap between bid and CPC as multiplicative noise on advertiser spend.
}
These metrics are reported at the campaign level and across \emph{segments}: combinations of categorical attributes such as placement, device, geography or user group.

Segment-level reporting is what makes diagnosis both possible and difficult.
The distribution of impressions across segments is shaped by interacting factors, including campaign configuration, competing campaigns, auction dynamics, and user behavior.
These factors also produce ordinary day-to-day metric noise.
Telling a real performance shift apart from this noise requires resolving three questions.
First, the cause is ambiguous: one metric movement can have several explanations (\emph{what}).
Second, the cause is localized: a segment-level effect can be diluted in the campaign aggregate and surface only under the right split (\emph{where}).
Third, the timing is unknown: gradual onsets are easily missed by simple period comparisons (\emph{when}).
Resolving all three at once is the core reasoning pattern \trace{} is designed to elicit.

\subsection{Data Generation Pipeline}\label{sec:pipeline}

For each \trace{} episode, the sampled intervention determines three hidden labels: the cause type, the affected segment slice, and the onset timing.

Generation proceeds in three stages: building a campaign population (Stage~1), sampling an intervention (Stage~2), and simulating the resulting metrics (Stage~3).

\paragraph{Stage 1: Campaign population.} 
We first construct a static universe of campaigns. 
Each is assigned a vertical category (e.g.\ electronics, fashion), a bidding strategy (e.g.\ search-heavy, acquisition), a baseline daily budget, and a baseline impression volume. 
A campaign is active in a sampled subset of its possible segments. 
Per-campaign variation in click-through and conversion rates is captured by a quality multiplier.

\paragraph{Stage 2: Intervention sampling.}

Each episode spans a contiguous window split into two equal parts: a \emph{baseline period}, in which the campaign runs normally, and an \emph{intervention period}, in which a sampled cause is active. 
We draw four components: 
the \emph{injected cause}, one of the twelve predefined cause types (\S\ref{sec:causes}); the \emph{driver slice}, which specifies the affected segments using one or two attributes, such as \texttt{placement=TOP\_OF\_SEARCH} or \texttt{geo=US} $\wedge$ \texttt{device=mobile}; the \emph{signal strength}, which sets the effect magnitude; and the \emph{onset profile}, which determines whether the effect appears immediately, after a delay, or gradually over time.


\paragraph{Stage 3: Metric simulation.}
The simulator renders daily metrics in three steps. 
First, it draws the campaign's total daily impressions from a baseline volume, a seasonality factor, and any volume change induced by active causes.
Second, it distributes these impressions across segments. 
The base distribution reflects the bidding strategy: a search-heavy campaign, for instance, weights search placements more heavily. 
Active causes then reweight this distribution, shifting impressions between segments. 
Third, it computes CTR, CVR, and CPC per segment from base rates, adjusted by the campaign's quality multiplier and by any cause effects applied to the driver slice. 
Clicks and orders are sampled from these rates, and spend follows from clicks and CPC.

Equation~\eqref{eq:sim_compact} summarizes the simulator. 
Active causes can affect the generated data through three channels: total impression volume, segment allocation, and per-segment rates.

\begin{equation}
  \begin{aligned}
  \text{impressions:}\quad
  N_{i,t}
  &\sim \mathrm{Poisson}\!\left(\mu_i \, \eta_t \, V_{i,t}\right), \\[3pt]
  \text{segment mix:}\quad
  \boldsymbol{\pi}_{i,t}
  &=
  \mathrm{Normalize}\!\left(\mathbf{w}_i \odot \mathbf{R}_{i,t}\right), \\
  \phantom{\text{segment mix:}\quad}
  \mathbf{N}_{i,t}
  &\sim \mathrm{Dirichlet\text{-}Multinomial}\!\left(
    N_{i,t}, \alpha \, \boldsymbol{\pi}_{i,t}
  \right), \\[3pt]
  \text{segment rates:}\quad
  \theta^{m}_{i,g,t}
  &=
  \bar{\theta}^{m}_{i,g}
  \cdot Q^{m}_{i}
  \cdot F^{m}_{i,g,t},
  \; \;
  m \in \{\mathrm{CTR}, \mathrm{CVR}, \mathrm{CPC}\}.
  \end{aligned}
  \label{eq:sim_compact}
\end{equation}

Here $i$ indexes campaigns, $t$ indexes days, and $g$ indexes segments.
The total impressions $N_{i,t}$ are sampled from the campaign baseline volume $\mu_i$, a seasonality factor $\eta_t$, and an intervention-induced volume multiplier $V_{i,t}$. 
Segment impressions $\mathbf{N}_{i,t}=(N_{i,g,t})_g$ are drawn from a multinomial distribution with probabilities $\boldsymbol{\pi}_{i,t}$, obtained by reweighting the campaign's baseline segment weights $\mathbf{w}_i$ by an intervention-induced segment-mix multiplier $\mathbf{R}_{i,t}$. 
The concentration parameter $\alpha$ controls day-to-day variability in segment allocation.
Finally, $\theta^m_{i,g,t}$ denotes the segment-level rate for metric $m \in \{\mathrm{CTR}, \mathrm{CVR}, \mathrm{CPC}\}$, with baseline rate $\bar{\theta}^m_{i,g}$, campaign quality multiplier $Q^m_i$, and intervention-induced rate multiplier $F^m_{i,g,t}$.

\subsection{Cause Taxonomy}\label{sec:causes}

\trace defines twelve cause types in four categories (Table~\ref{tab:cause_signatures}), organized by \emph{where} the causal signal appears. The categories require different diagnostic strategies, so no fixed query procedure is sufficient.

\emph{Campaign-wide causes} produce effects visible in aggregate campaign metrics.

\emph{Segment-mix causes} redistribute impressions across segments while leaving campaign totals roughly unchanged. 
Their signature is a change in traffic composition across periods. 

\emph{Segment-specific causes} alter rates inside a localized \emph{driver slice}, leaving other segments unaffected. 
These effects can be diluted in aggregate metrics and surface only under the right split. 
This category is the hardest: the agent must both localize the driver slice and match the multi-metric signature that distinguishes one cause from another.

Finally, \emph{no-signal} episodes inject no cause at all: every fluctuation is stochastic noise or a confounder. 
These episodes are not trivial because confounders, such as seasonality, can appear regardless of the cause. 
They test whether an agent abstains when evidence is insufficient, penalizing models that default to plausible-sounding attributions.

\begin{table}[t]
  \centering
  \caption{Root-cause signatures, grouped by \emph{signal level}.}
  \label{tab:cause_signatures}
  \small
  \setlength{\tabcolsep}{5pt}
  \renewcommand{\arraystretch}{1.2}
  \begin{tabular}{@{}p{4.0cm}p{5.0cm}p{1.7cm}@{}}
    \toprule
    \textbf{Cause} & \makecell[l]{\textbf{Metric}\\\textbf{Signature}}
    & \makecell[l]{\textbf{Driver}\\\textbf{Dimension}} \\
    \midrule
    \multicolumn{3}{@{}l@{}}{\textbf{Campaign-wide}} \\
    \addlinespace[2pt]
    \textsc{bid increase}        & CPC$\uparrow$, CTR$\downarrow$\,(mild)                          & --- \\
    \textsc{budget cap}          & Impr$\downarrow$, clicks$\downarrow$, orders$\downarrow$; rates$\approx$ & --- \\
    \textsc{page degradation} & CVR$\downarrow$, CTR$\approx$, orders$\downarrow$                        & --- \\
    \textsc{out of stock}       & stock$\downarrow$, CVR$\downarrow$, orders$\downarrow$         & --- \\
    \midrule
    \multicolumn{3}{@{}l@{}}{\textbf{Segment-mix}} \\
    \addlinespace[2pt]
    \textsc{placement shift}      & segment share $\updownarrow$    & placement \\
    \textsc{targeting broadening} & Impr$\uparrow$, CTR$\downarrow$, CVR$\downarrow$ & audience  \\
    \textsc{targeting narrowing}  & Impr$\downarrow$, CTR$\uparrow$, CVR$\uparrow$   & audience  \\
    \midrule
    \multicolumn{3}{@{}l@{}}{\textbf{Segment-specific}} \\
    \addlinespace[2pt]
    \textsc{creative fatigue}      & CTR$\downarrow$ (gradual), Impr$\approx$, CVR$\approx$ & \makecell[l]{placement/\\device/geo} \\
    \textsc{competitive pressure}  & CPC$\uparrow$, CTR$\downarrow$             & geo      \\
    \textsc{ad quality drop}     & Impr$\downarrow$, CPC$\uparrow$, CTR$\downarrow$ & device   \\
    \textsc{audience saturation}   & CTR$\downarrow$ (gradual), CVR$\downarrow$\,(mild) & audience \\
    \midrule
    \multicolumn{3}{@{}l@{}}{\textbf{No signal}} \\
    \addlinespace[2pt]
    \textsc{no signal}             & all $|\Delta|$ within noise                & --- \\
    \bottomrule
  \end{tabular}
\end{table}

\paragraph{Complexity axes.}
Episode difficulty is controlled by two cause-conditioned axes.
\emph{Dimensional complexity} controls \emph{where}: the signal's driver slice may filter on one attribute or, in harder cases, the intersection of two.
\emph{Temporal complexity} controls \emph{when}: the signal's onset profiles can be immediate, delayed, or ramping.
Together with the cause taxonomy, these axes test whether the agent can identify what changed and where the signal appears under different temporal patterns.

\subsection{Database and Task Interface}\label{sec:task}

For each episode, the agent receives query access to \emph{fact tables} that contain daily metrics at the campaign and segment level. 
The ground-truth labels are stored in separate tables that are used only for verification, scoring and training reward, and they are never exposed to the agent.
The full schema is given in Appendix~\ref{app:schema}.

The agent receives a system prompt specifying its role, the candidate cause types, the available tables, and the required final-answer format.
The user prompt describes the observed performance change for a specific campaign.
The agent interacts through a Python tool backed by a sandboxed SQL engine, with state persisted across calls so that later analyses can build on earlier queries.

The agent's final answer contains a target cause and supporting evidence.
Segment-specific causes additionally require an affected segment slice, such as \texttt{placement=TOP\_OF\_SEARCH}. 
In harder episodes, the slice may be an intersection of two attributes, such as \texttt{placement=TOP\_OF\_SEARCH, geo=US}. 
Campaign-wide, segment-mix, and no-signal episodes require no separate segment field.

\subsection{Solvability and Verification}\label{sec:verifier}

A useful benchmark here must meet three competing requirements.
Episodes must be \emph{solvable}: the injected signal must be recoverable from the agent-visible data. 
They must be \emph{non-trivial}: simple signature matching should not be sufficient.
And they must be \emph{realistic}: metrics should covary and confounders should be present, as in operational data. 
These requirements are in tension. A signal weak enough to avoid trivial detection may become unrecoverable, while a signal strong enough to guarantee recovery may collapse the task into pattern matching.

\trace{} resolves this tension by separating generation from acceptance. 
The simulator first generates diverse episodes with varying causes, complexity axes, noise, and confounders.
An oracle verifier then admits only episodes that remain recoverable from the agent-visible data.
The verifier knows the ground-truth cause but reads only the agent-visible data. 
It does not solve the episode. Instead, it verifies that the injected signal is detectable, stronger than the strongest confounder, and distinguishable from competing causes. 
Accepted episodes therefore retain realistic ambiguity while providing objective labels for scoring and reward construction. 

\section{RL with Synthesized Rewards}\label{sec:training}

The \trace{} environment turns each accepted episode into an agent-visible diagnostic task paired with a hidden oracle label. 
The task defines the data observed by the agent, while the label records the injected intervention. 
Comparing a trajectory's final attribution with this label produces an objective synthesized reward without human annotation or an LLM judge. 
This setup enables us to train an open-weight diagnostic agent under the same tool-use interface used for evaluation.

\subsection{Reward Design}\label{sec:rl}

We optimize the policy with a GRPO objective~\citep{shao2024deepseekmath} using stabilization refinements
from recent RLVR recipes~\citep{yu2026dapo}.
For each training episode, the policy samples a group of trajectories and receives a synthesized reward computed against the hidden oracle label.

The reward combines three terms: a graded attribution reward, a binary full-attribution reward, and a small formatting reward:
\begin{equation}
  r =
  w_{\mathrm{attr}} r_{\mathrm{attr}}
  + w_{\mathrm{full}} r_{\mathrm{full}}
  + w_{\mathrm{fmt}} r_{\mathrm{fmt}} .
  \label{eq:train_reward}
\end{equation}
The weights are nonnegative and sum to one, so $r\in[0,1]$.

The attribution reward decomposes into cause correctness and slice specification:
\begin{equation}
  r_{\mathrm{attr}}
  =
  \mathbf{1}\{\hat{c}=c^\star\}\, r_{\mathrm{slice}},
  \label{eq:attr_reward}
\end{equation}
where $\hat{c}$ is the predicted cause, $c^\star$ is the oracle cause, and $r_{\mathrm{slice}}$ scores slice specification when an affected segment slice is required.
The reward term $r_{\mathrm{attr}}$ gates attribution by cause correctness: a wrong cause receives zero attribution reward.

For segment-specific episodes, let $\widehat{Z}$ and $Z^\star$ denote the predicted and oracle driver slices, represented as sets of dimension--value pairs.
We measure their agreement using Jaccard similarity:
$
J(\widehat{Z}, Z^\star)
=
\frac{|\widehat{Z}\cap Z^\star|}
     {|\widehat{Z}\cup Z^\star|}
$.
The graded slice reward is
\begin{equation}
r_{\mathrm{slice}}
=
\alpha + (1-\alpha)J(\widehat{Z}, Z^\star),
\label{eq:slice_reward}
\end{equation}
which provides partial credit for identifying some, but not all, of the affected dimensions. 
For campaign-wide, segment-mix, and no-signal episodes, where no driver slice is required, we set $r_{\mathrm{slice}}=1$.


The graded attribution reward assigns substantial partial credit to an incomplete multi-dimensional slice. 
To provide a stronger incentive for recovering the complete driver slice, we introduce a binary full-attribution reward $r_{\mathrm{full}}\in\{0,1\}$. 
We set $r_{\mathrm{full}}= \mathbf{1} \{r_{\mathrm{attr}}=1\}$, 
which equals one only when the predicted cause is correct and the predicted driver slice exactly matches the oracle slice.
Thus, $r_{\mathrm{attr}}$ provides dense partial credit, whereas $r_{\mathrm{full}}$ rewards a complete attribution.

The last term $r_{\mathrm{fmt}}\in\{0,1\}$ is a format reward indicating  a parseable final answer under the required schema. 
Parse failures results in $r_{\mathrm{attr}}=0$.
In our main results, we set $(w_{\mathrm{attr}}, w_{\mathrm{full}}, w_{\mathrm{fmt}}) = (0.65, 0.30, 0.05)$ and $\alpha=0.5$. 

For reward computation, we parse the final answer to extract the decision fields: the root cause and, when required, the affected segment slice. 
Evidence is collected for interpretability and error analysis, but it does not affect the reward. 
This keeps training and evaluation focused on the same attribution target.

\subsection{Training Setup}\label{sec:training_details}
Our trained agent is based on \textbf{Qwen3.5-35B-A3B}, a mixture-of-experts open-weight model with 35B total parameters and 3B active parameters.
The RL dataset contains 5,000 oracle-verified episodes, split into 4,472 training and 528 validation tasks, with stratification by cause, signal level, and 1D/2D driver slice complexity. 
The evaluation dataset contains 235 held-out tasks that are disjoint from the training and validation tasks at episode, campaign, and intervention levels.


Training and evaluation use the same multi-turn tool interface.
The agent executes Python containing SQL queries over agent-visible fact tables and return a structured final answer for cause attribution. 

We compare two RL initializations, the base model and an SFT warm start checkpoint, and study the impact of reward shape and KL regularization with ablation experiments in \S\ref{sec:ablations}.
Full optimizer, rollout, tool-execution, hardware, and parallelism details are provided in Appendix~\ref{app:training_details}.

\subsection{Supervised Warm Start}\label{sec:sft}

The supervised fine-tuning (SFT) as a warm-up stage before RL teaches the required answer format and multi-turn tool-use pattern, and exposes the model to valid diagnostic trajectories. 
We construct 1,200 SFT examples by rejection-sampling complete teacher trajectories and retaining only those whose final cause and driver slice match the oracle label. 
Each example contains the task prompt, tool calls, tool outputs, and final structured answer.


Candidate teacher trajectories are generated by Claude Opus~4.8.
For difficult segment-specific episodes, we also provide the teacher model with lightweight hints about the root-cause signature, as a way of improving rejection-sampling efficiency. 
These hints are not included in the retained task prompts, RL or evaluation prompts. 

\section{Experiments}\label{sec:experiments}
We use \trace{} to evaluate whether synthesized rewards can train a stronger tool-using diagnostic agent than prompting alone.
Our experiments ask four questions.
Is the held-out \trace{} benchmark challenging for strong frontier models? 
Does RL with synthesized rewards improve full attribution when initialized from either the base model or an SFT warm start?
How do reward design and policy initialization affect the gains from RL?
Finally, how does post-training change tool-call efficiency?

\subsection{Experimental Setup}\label{sec:eval_setup}

\paragraph{Evaluation dataset.} 
All reported evaluations use the same held-out 235-episode \trace{} test dataset.
The dataset is deliberately enriched for segment-specific cases: 
164 episodes require identifying a driver slice, including 49 whose oracle slice intersects two dimensions.
This composition emphasizes exact attribution under fine-grained segmentation rather
than only campaign-level cause identification.


\paragraph{Models.}
We evaluate four variants of Qwen3.5-35B-A3B: the base model, an SFT model, an RL model initialized from the base model, and an RL model initialized from the SFT checkpoint.
We compare these variants with Qwen3.5-122B-A10B, Claude Opus~5, Claude Sonnet~5, GPT-5.5, and GPT-5.6 Sol baselines.

\paragraph{Metrics.}
We evaluate diagnostic correctness using the same cause and driver-slice criteria used by the attribution rewards in Section~\ref{sec:rl}.
Let $C=\mathbf{1}\{\hat{c}=c^\star\}$ denote root-cause correctness.
For segment-specific episodes, driver-slice agreement is measured by the Jaccard similarity $J(\widehat{Z},Z^\star)$.
A trajectory achieves \emph{full attribution} when $r_{\mathrm{full}}=1$:
the cause is correct and, when a driver slice is required, the predicted slice exactly matches the oracle slice.

We report \textbf{Cause@$k$}, the probability that at least one of $k$ sampled trajectories identifies the correct cause, and \textbf{FullAttr@$k$}, the probability for full attribution.
We additionally report mean driver-slice Jaccard similarity conditioned on a correct cause for segment-specific episodes, no-signal accuracy, decision-parse rate, and mean executed tool calls.



\subsection{Main Results}\label{sec:main_results}
Table~\ref{tab:main_results} summarizes our main experiment results and shows three major findings. 

First, \trace{} is \textbf{hard and unsaturated}:  the strongest prompted baseline,
Claude Opus~5, achieves only $0.686$ FullAttr@1, with substantial remaining errors on
segment-specific episodes (Section~\ref{sec:difficulty}).

Second, \textbf{RL with synthesized rewards is additive to SFT}. SFT alone reaches
$0.637$ FullAttr@1, while SFT$\rightarrow$RL improves it by 12.0 percentage points to
$\mathbf{0.757}$, surpassing all prompted baselines.

Third, \textbf{post-training can outweigh prompted model scale}. 
Within the Qwen3.5 family, the post-trained 35B model substantially outperforms the prompted 122B model ($0.757$ versus $0.283$) and also exceeds every evaluated closed-source baseline.
This suggests that \trace{} rewards a learnable diagnostic procedure that prompting and additional model scale alone do not reliably elicit.

RL initialized directly from the 35B base also improves FullAttr@1 substantially ($0.159\rightarrow0.434$), but remains below both SFT and SFT$\rightarrow$RL.
Its decision-parse rate is also only $0.68$, compared with $1.00$ for both SFT-initialized models, suggesting that the supervised warm start helps the policy produce scoreable decisions as well as improve attribution.

\begin{table}[t]
  \centering
  \caption{Performance on the 235-episode held-out \trace{} test set under reward-aligned decision parsing.
Cause@1 evaluates root-cause identification.
FullAttr@$k$ requires the correct cause and, for segment-specific episodes, an exact driver-slice match.
Jaccard is mean Jaccard similarity on cause-correct segment-specific trajectories; 
No-Signal is accuracy on no-signal episodes; 
Parsed is the decision-extraction rate.}
  \label{tab:main_results}
  \small
  \setlength{\tabcolsep}{3.5pt}
  \begin{tabular}{lcccccc}
    \toprule
    \textbf{Model} & \textbf{Cause@1} & \textbf{Jaccard}
    & \textbf{FullAttr@1} & \textbf{FullAttr@5} 
    & \textbf{No-Signal} & \textbf{Parsed} \\
    \midrule
    Qwen3.5-35B     & 0.184 & 0.596 & 0.159 & 0.370 & 0.41 & 0.53 \\
    \quad + SFT                  & 0.685 & 0.958 & 0.637 & 0.762 & 0.76 & 1.00 \\
    \quad + RL        & 0.471 & 0.913 & 0.434 & 0.711 & 0.28 & 0.68 \\
    \quad + SFT $\rightarrow$ RL          & \textbf{0.823} & 0.928 & \textbf{0.757} & \textbf{0.851} & 0.49 & 1.00 \\
    \midrule
    Qwen3.5-122B     & 0.296 & 0.842 & 0.283 & 0.434 & 0.78 & 0.88 \\
    Claude Opus 5                & 0.764 & 0.913 & 0.686 & 0.809 & 0.87 & 0.99 \\
    GPT-5.6 Sol                  & 0.635 & 0.852 & 0.565 & 0.719 & 0.30 & 0.99 \\
    GPT-5.5                      & 0.581 & 0.893 & 0.524 & 0.643 & 0.61 & 0.99 \\
    Claude Sonnet 5              & 0.472 & 0.899 & 0.438 & 0.607 & 0.85 & 1.00 \\
    \bottomrule
  \end{tabular}
\end{table}

\subsection{Training Ablations}\label{sec:ablations}

Table~\ref{tab:ablations} examines the effects of the full-attribution reward, policy initialization, and KL regularization.

\textbf{The full-attribution reward primarily improves difficult slice attribution.}
With SFT initialization, RL using only the graded attribution reward reaches $0.596$ FullAttr@1 and obtains no exact matches on episodes with two-dimensional driver slices.
Adding $r_{\mathrm{full}}$ increases overall FullAttr@1 to $0.757$, with gains from $0.67$ to $0.92$ on one-dimensional slices and from $0.00$ to $0.27$ on two-dimensional slices.
This pattern suggests that an explicit reward for complete attribution is especially important when success requires recovering multiple driver dimensions.

\textbf{SFT initialization remains important.}
Under the same reward containing $r_{\mathrm{full}}$, RL initialized from the base model reaches $0.434$ FullAttr@1, compared with $0.757$ when initialized from SFT.
Thus, RL improves the base policy, but does not recover the performance of the supervised warm start under the matched training recipe.

\textbf{KL regularization does not improve the observed result.}
Adding a KL penalty to the SFT reference reduces FullAttr@1 from $0.757$ to $0.724$.
Performance on two-dimensional slices decreases from $0.27$ to $0.16$, while one-dimensional performance remains unchanged at $0.92$.
No-signal accuracy also decreases from $0.49$ to $0.30$.

\begin{table}[t]
  \centering
  \caption{Training ablations on the held-out \trace{} test set. All columns report
  FullAttr@1, either overall or on the indicated episode subset. The
  full-attribution condition adds the binary $r_{\mathrm{full}}$ term to the graded
  attribution reward.}
  \label{tab:ablations}
  \small
  \begin{tabular}{lcccc}
    \toprule
    \textbf{Training condition} & \textbf{Overall} & \textbf{1D Slice}
    & \textbf{2D Slice} & \textbf{No Signal} \\
    \midrule
    SFT only                                      & 0.637 & 0.71 & 0.04 & 0.76 \\
    SFT $\rightarrow$ RL, graded reward           & 0.596 & 0.67 & 0.00 & 0.64 \\
    SFT $\rightarrow$ RL, + full-attribution term & \textbf{0.757}
                                                  & \textbf{0.92}
                                                  & \textbf{0.27} & 0.49 \\
    SFT $\rightarrow$ RL, + full-attribution term + KL
                                                  & 0.724 & 0.92 & 0.16 & 0.30 \\
    RL from base, + full-attribution term         & 0.434 & 0.51 & 0.03 & 0.28 \\
    \bottomrule
  \end{tabular}
\end{table}

\subsection{Performance Breakdown}\label{sec:difficulty}

Table~\ref{tab:breakdown} decomposes FullAttr@1 by cause scope and driver-slice
dimensionality. 
SFT and SFT$\rightarrow$RL approach ceiling on campaign-wide and segment-mix causes. 
Post-training provides its largest gain over SFT on one-dimensional segment-specific episodes, improving FullAttr@1 from $0.71$ to $0.92$.

Two-dimensional attribution remains the principal challenge. 
SFT$\rightarrow$RL improves FullAttr@1 from $0.04$ to $0.27$ on these episodes, but still performs below Claude Opus~5 at $0.33$. 
RL also reduces no-signal accuracy from $0.76$ after SFT to $0.49$, revealing a trade-off between stronger attribution and avoiding false-positive diagnoses.

\begin{table}[t]
  \centering
  \caption{FullAttr@1 by episode type. $N$ is the number of test episodes in each
  subset. Two-dimensional driver slices remain the most difficult attribution setting.}
  \label{tab:breakdown}
  \small
  \begin{tabular}{lccccc}
    \toprule
    \textbf{Subset} & \textbf{$N$} & \textbf{Base} & \textbf{SFT}
    & \textbf{SFT$\rightarrow$RL} & \textbf{Opus 5} \\
    \midrule
    Campaign-wide causes         & 30  & 0.46 & \textbf{0.99} & 0.95 & 0.95 \\
    Segment-mix causes           & 21  & 0.48 & 0.98 & \textbf{1.00} & 0.98 \\
    Segment-specific, 1D slice   & 115 & 0.04 & 0.71 & \textbf{0.92} & 0.68 \\
    Segment-specific, 2D slice   & 49  & 0.01 & 0.04 & 0.27 & \textbf{0.33} \\
    No-signal episodes           & 20  & 0.41 & 0.76 & 0.49 & \textbf{0.87} \\
    \bottomrule
  \end{tabular}
\end{table}

To examine two-dimensional errors more closely, we analyze all five saved trajectories per episode. 
Among cause-correct trajectories whose oracle slice contains a \texttt{device} dimension, SFT$\rightarrow$RL recovers the correct device--value pair in $66/94$ cases, compared with $14/39$ for SFT and $12/31$ for RL from the base.
Thus, the strongest trained model is substantially better at recovering this second dimension, although incomplete slices remain common.

The model also tends to over-attribute changes when no signal is present.
SFT$\rightarrow$RL predicts \textsc{no\_signal} in only $49/100$ saved no-signal trajectories, compared with $76/100$ after SFT.
Another recurring ambiguity occurs between segment-specific causes with similar observable signatures. 
Of the 46 SFT$\rightarrow$RL failures to identify \textsc{ad\_quality\_degradation}, 24 predict \textsc{competitive\_pressure}.
These results suggest that further gains require both more reliable multi-dimensional slice recovery and better calibration among related causes.

\subsection{Tool-Call Efficiency}\label{sec:tool_efficiency}

\begin{figure}[t]
  \centering
  \includegraphics[width=0.8\linewidth]{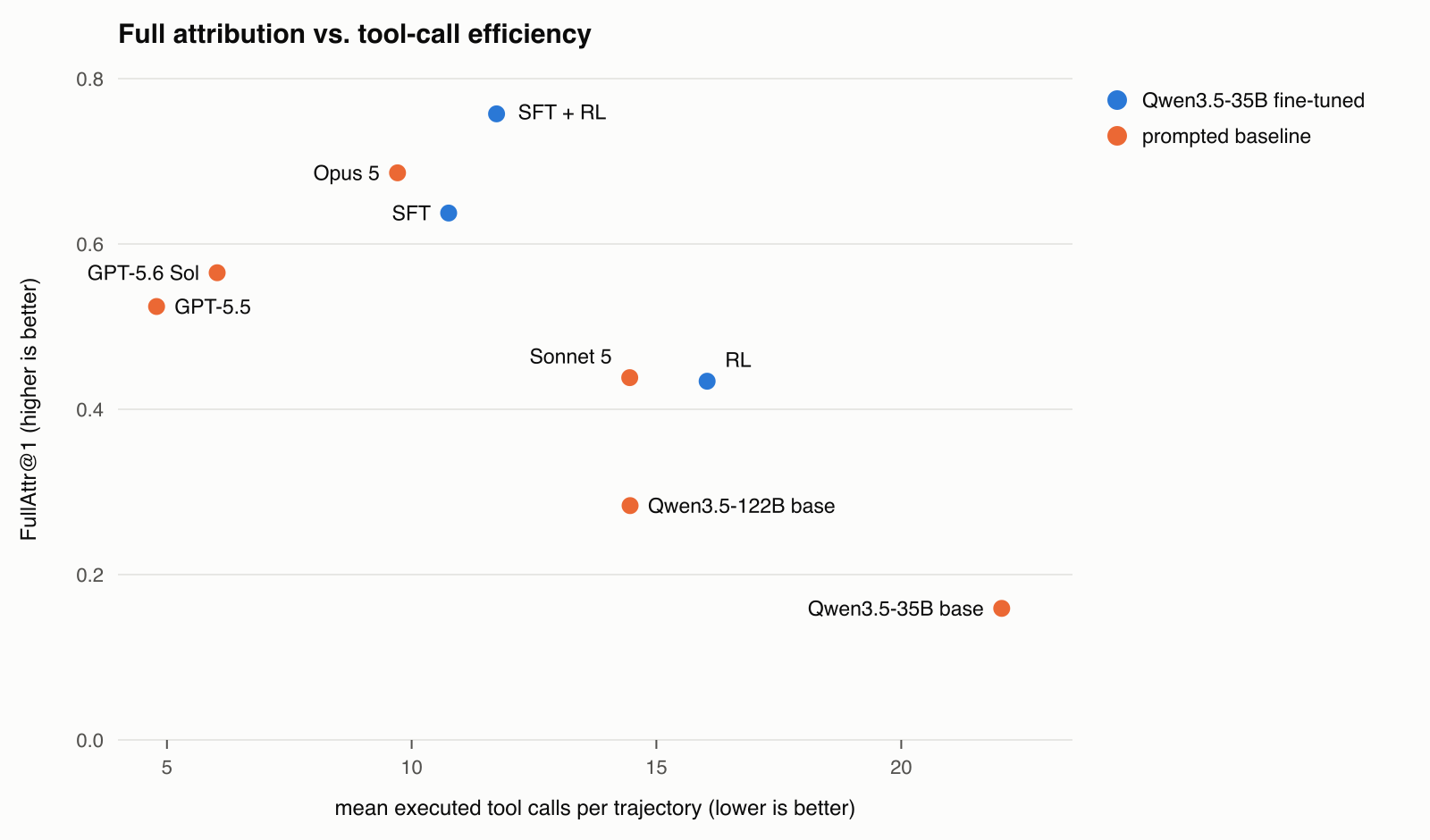}
  \caption{FullAttr@1 versus mean executed tool calls per evaluation trajectory.
  Each trajectory entry corresponds to one executed Python-tool call, including
  unsuccessful calls. Blue points denote fine-tuned Qwen3.5-35B variants, and orange
  points denote prompted baselines. Better performance lies toward the upper left.}
  \label{fig:tool_efficiency}
\end{figure}

Figure~\ref{fig:tool_efficiency} compares FullAttr@1 with the mean number of executed tool calls per evaluation trajectory. 
SFT simultaneously improves attribution and reduces mean tool use relative to the prompted 35B base, from $22.05$ to $10.75$ calls.
RL after SFT increases mean tool use by only $0.98$ calls, to $11.73$, while improving FullAttr@1 by 12.0 percentage points. 
In contrast, RL from the base averages $16.04$ calls while remaining less accurate than SFT. 
These results indicate that the post-training gains are not explained by simply making more tool calls. 
They also show that supervised warm start teaches a more effective investigation-and-stopping procedure, rather than merely encouraging additional exploration.


\section{Conclusion}\label{sec:conclusion}

We introduced a simulator--oracle--RL approach for training diagnostic-reasoning agents when naturally occurring verifiers are scarce. 
A controlled simulator samples a hidden intervention, generates the data resulting from that intervention, and retains it as an oracle label. 
This construction makes the final attribution objectively verifiable without removing the noise, confounders, multi-table evidence, and exploratory tool use that make the diagnostic task difficult. 
We instantiate this approach in \trace{}, a generative digital-advertising environment containing campaign-wide, segment-mix, and segment-specific root causes.

On the held-out \trace{} benchmark, the strongest prompted baseline reaches $0.686$ FullAttr@1. 
SFT raises Qwen3.5-35B-A3B from $0.159$ to $0.637$, and subsequent RL with synthesized rewards further improves it to $0.757$, surpassing every evaluated prompted baseline, including Qwen3.5-122B-A10B. 
This result provides evidence that, for this diagnostic setting, the binding constraint is access to an effective post-training signal---including a scalable, objective reward---rather than model scale alone.


Several directions could extend this approach. 
Richer agent harnesses could provide explicit planning, memory, adaptive tool selection, and improved stopping mechanisms, allowing the policy to conduct longer and more structured investigations. 
The simulator--oracle construction could also be expanded to broader intervention families, held-out root causes, and domains such as software operations and data-quality diagnosis, where failures can be injected and their downstream effects observed.
Finally, simulator control enables systematic curricula over noise, confounding, evidence availability, and attribution complexity, providing a way to study which diagnostic procedures transfer across environments.

More broadly, our results demonstrate a route to constructing objective training signals for otherwise ambiguous diagnostic tasks: when a domain admits a controllable generative model, the intervention that creates a difficult problem can also provide its verifier.

\bibliographystyle{plainnat}
\bibliography{reference}

\appendix

\section{Database and Output Schemas}\label{app:schema}

\subsection{Agent-Visible Fact Tables}

Each episode provides access to a DuckDB fact store containing the four tables summarized in
Table~\ref{tab:database_schema}. Rows are keyed by campaign identifiers, and the task prompt specifies the
campaign to investigate. The agent can access these tables only through its Python/SQL tool; no
ground-truth field is included in the tool-visible database.

\begin{table}[h]
  \centering
  \caption{Agent-visible database schema. All tables include a date field \texttt{ds} and campaign
  identifiers as appropriate.}
  \label{tab:database_schema}
  \small
  \begin{tabular}{p{0.22\linewidth}p{0.18\linewidth}p{0.50\linewidth}}
    \toprule
    \textbf{Table} & \textbf{Grain} & \textbf{Fields} \\
    \midrule
    \texttt{daily\_campaign}
      & Campaign-day
      & Campaign identifiers and category; budget, spend, impressions, clicks, orders, sales,
        page views, and brand searches. \\
    \texttt{segment\_daily}
      & Segment-day
      & Campaign identifiers and category; ad product, targeting type, match type, placement, price band, audience, device, and geography; delivery, engagement, and
        conversion metrics. \\
    \texttt{budget\_log}
      & Campaign-day
      & Advertiser and campaign identifiers, budget, and spend. \\
    \texttt{inventory}
      & Campaign-day
      & Campaign identifiers and category, together with the fraction of advertised products that are in stock. \\
    \bottomrule
  \end{tabular}
\end{table}

\subsection{Hidden Oracle Tables}

The generation database additionally contains three oracle-only tables. \texttt{gt\_episode} records the
injected root cause, affected segment assignment, intervention strength, temporal pattern, and number of
affected dimensions. \texttt{gt\_evidence} stores post-computed evidence associated with the intervention,
including its source table, metric, segment, direction, and change relative to the baseline period.
\texttt{gt\_validation} records the oracle verifier's detectability and confounder-elimination checks.
These tables are retained for dataset construction and scoring but are excluded when the agent-visible
database is created.

The required final answer contains a \texttt{decision} object with a non-empty \texttt{root\_cause} and
an optional \texttt{driver\_segments} list, an \texttt{evidence} list, and a textual \texttt{explanation}.
For campaign-wide, segment-mix, and no-signal episodes, \texttt{driver\_segments} is omitted or null.
For segment-specific episodes, it contains one or more dimension--value assignments.

\subsection{Oracle Verification}\label{app:verifier}

After simulation, the oracle verifier compares the episode window with an equal-length preceding baseline
window. It first checks that the expected signature is present at the appropriate level: campaign metrics
for campaign-wide causes, impression-share changes for segment-mix causes, and changes within the injected
segment for segment-specific causes. For delayed and ramping interventions, it also tests the second half
of the episode window so that a recoverable late-onset signal is not rejected solely because it is diluted
in the full-window average. Segment-specific signals must additionally exceed their own baseline temporal
variation, using a minimum signal-to-noise ratio of 1.

The verifier then tests each alternative cause against the same agent-visible data. An alternative is
eliminated using differences in signal level, source table, affected dimension, primary metric, or
companion-metric direction. Episodes are rejected when the injected signal is absent or when a matching
alternative cannot be eliminated. No-signal episodes are retained only when observed fluctuations do not
realize another cause's signature. Thus, acceptance uses the hidden intervention to define what must be
verified, but all detectability and distinguishability checks operate on the same fact tables available to
the agent.

\section{Training Details}\label{app:training_details}

\subsection{Training Setup}

We train Qwen3.5-35B-A3B using a \texttt{slime} fork with Megatron-LM for optimization,
SGLang for rollout inference, and Ray for orchestration. RL uses an asynchronous pipeline in which rollout
generation for the next step overlaps policy optimization, with updated weights synchronized from
Megatron-LM to SGLang after each optimizer step. Training uses bfloat16 parameters with float32 gradient
accumulation, softmax, and router computation.

\subsection{Supervised Fine-Tuning}

The SFT stage trains on 1,200 oracle-filtered teacher trajectories for three epochs. We use a learning rate
of $10^{-5}$ with cosine decay and apply the language-model loss only to assistant turns under the
Qwen3.5 chat template. Each example retains the complete interaction, including prompts, tool calls, tool
outputs, and the final structured answer. 

\subsection{Reinforcement Learning}

Table~\ref{tab:rl_hyperparameters} gives the shared GRPO configuration. Each optimizer step draws 32
prompts and samples eight trajectories per prompt. The agent has a 32,768-token context window, a
30-turn backstop, and a maximum of 2,048 generated tokens per turn. Training rollouts use temperature
$1.0$ to maintain within-group exploration.

\begin{table}[h]
  \centering
  \caption{Shared hyperparameters for the GRPO runs.}
  \label{tab:rl_hyperparameters}
  \small
  \begin{tabular}{ll}
    \toprule
    \textbf{Hyperparameter} & \textbf{Value} \\
    \midrule
    Training set & 4,472 prompts \\
    Validation set & 528 prompts \\
    Optimizer steps & 300 (approximately two epochs) \\
    Prompts per step & 32 \\
    Samples per prompt & 8 \\
    Effective trajectories per step & 256 \\
    Optimizer & AdamW, $\beta=(0.9,0.98)$ \\
    Learning rate & $10^{-6}$, constant \\
    Weight decay & 0.1 \\
    PPO clipping range & $[0.8,1.28]$ \\
    Gradient clipping & 1.0 \\
    Entropy coefficient & 0 \\
    Rollout temperature & 1.0 \\
    Maximum turns & 30 \\
    Context length & 32,768 tokens \\
    Maximum generation per turn & 2,048 tokens \\
    Validation frequency & Every 20 optimizer steps \\
    Validation sampling & Five trajectories per prompt at temperature 0.6 \\
    \bottomrule
  \end{tabular}
\end{table}

The main RL condition uses the reward weights reported in Section~\ref{sec:rl}. The KL ablation adds a
penalty to the SFT reference policy with coefficient $0.005$, using the low-variance k3 estimator. All RL
conditions use the same data split, rollout group size, optimizer settings, and nominal 300-step budget.
The validation split is stratified by root cause, signal level, and one- versus two-dimensional segment
attribution and is disjoint from both the RL training prompts and SFT demonstrations.

\section{Additional Evaluation Details}\label{app:additional_eval}

\subsection{Evaluation Protocol}\label{app:evaluation_protocol}

All models receive the same task specification, candidate-cause definitions, final-answer contract, and
Python/SQL tool interface over the agent-visible fact store. The Python state persists across calls within
an episode. Tool calls have a 60-second execution timeout, and tool outputs are truncated after 8,000
characters. A final-turn nudge asks the model to return its answer before exhausting the interaction
budget.

Open-weight evaluations sample five trajectories per task at temperature $0.6$ with a 60-step backstop.
API-model evaluations use provider-supported stochastic sampling without an explicit temperature and a
30-turn backstop. The reported Claude and GPT baselines use the \texttt{xhigh} reasoning-effort setting.
Claude models receive a maximum output-token budget of 32,768, while GPT models receive 16,384 output
tokens. Every reported FullAttr@5 value is computed from five distinct samples per task. The interaction
limits do not bind for the trained or frontier models; one Qwen3.5-35B base-model trajectory is truncated.

For $n$ sampled trajectories containing $c$ successful trajectories, Cause@$k$ and FullAttr@$k$ use the
standard unbiased estimator
\begin{equation}
  1-\frac{\binom{n-c}{k}}{\binom{n}{k}}.
\end{equation}
Cause@$k$ treats root-cause correctness as success, whereas FullAttr@$k$ additionally requires an exact
segment assignment when the episode is segment-specific. At $k=1$, the estimator is the mean
single-trajectory success probability over the test tasks.

\subsection{Answer-Parser Sensitivity}\label{app:parser_sensitivity}

For the primary evaluation, we parse each model's final answer to extract the decision fields needed to
assess attribution: a non-empty root cause and any predicted segment assignment. Malformed report-only
evidence, an omitted explanation, or benign extra JSON keys do not invalidate an otherwise scoreable
decision. A malformed segment assignment is treated as absent and therefore cannot earn full attribution
on a segment-specific episode. We report the decision-parse rate separately from diagnostic accuracy.

As a sensitivity analysis, we also use an exact-schema parser that requires the complete prescribed output
structure. Table~\ref{tab:parser_sensitivity} reports a paired re-scoring of the same saved trajectories
under both parsers. The parser choice has no effect on the SFT or SFT$\rightarrow$RL results, but
exact-schema parsing underestimates FullAttr@1 when a model produces a semantically scoreable decision
with malformed or missing report-only fields.

\begin{table}[h]
  \centering
  \caption{FullAttr@1 obtained by re-scoring the same trajectories with an exact-schema parser and the
  decision parser used for the main results.}
  \label{tab:parser_sensitivity}
  \small
  \begin{tabular}{lcc}
    \toprule
    \textbf{Model} & \textbf{Exact Schema} & \textbf{Decision Parser} \\
    \midrule
    Qwen3.5-35B                  & 0.140 & 0.159 \\
    \quad + SFT                  & 0.637 & 0.637 \\
    \quad + RL                   & 0.317 & 0.434 \\
    \quad + SFT $\rightarrow$ RL & 0.757 & 0.757 \\
    Claude Opus 5                & 0.665 & 0.686 \\
    \bottomrule
  \end{tabular}
\end{table}

\end{document}